\documentclass{article}
\pdfoutput=1

\usepackage[nonatbib,preprint]{neurips_2021}

\title{Scaling Ensemble Distribution Distillation to Many Classes with Proxy Targets}

\author{%
  Max Ryabinin\thanks{Equal contribution.} \\
  Yandex, HSE University\\
  Moscow, Russia \\
  \texttt{mryabinin0@gmail.com} \\
  \And 
   Andrey Malinin
   \footnotemark[1] \\
   Yandex, HSE University \\
   Moscow, Russia \\
   \texttt{am969@yandex-team.ru} \\
  \And
   Mark Gales \\
   University of Cambridge \\
   Cambridge, United Kingdom \\
   \texttt{mjfg@eng.cam.ac.uk} \\
}

\usepackage{amsmath,amsfonts,bm}

\def\eqref#1{equation~\ref{#1}}

\def\1{\bm{1}}

\DeclareMathAlphabet{\mathsfit}{\encodingdefault}{\sfdefault}{m}{sl}
\SetMathAlphabet{\mathsfit}{bold}{\encodingdefault}{\sfdefault}{bx}{n}

\usepackage{hyperref}
\usepackage{url}

\usepackage[utf8]{inputenc} 
\usepackage[T1]{fontenc}    
\usepackage{hyperref}       
\usepackage{url}            
\usepackage{booktabs}       
\usepackage{amsfonts}       
\usepackage{nicefrac}       
\usepackage{microtype}      
\usepackage{xcolor}         

\usepackage{subfigure}
\usepackage[titletoc,title]{appendix}
\usepackage{mathtools}
\usepackage{subfiles}
\usepackage{latexsym}
\usepackage{multirow}
\usepackage{empheq}
\usepackage{bm}
\usepackage{graphicx}

\usepackage[normalem]{ulem}
\useunder{\uline}{\ul}{}
\usepackage{xspace}
\usepackage{siunitx}    
\newcommand{\Endd}{EnD$^2$\xspace}

\begin{document}

\maketitle

\begin{abstract}
    Ensembles of machine learning models yield improved system performance as well as robust and interpretable uncertainty estimates; however, their inference costs may often be prohibitively high.
\emph{Ensemble Distribution Distillation} is an approach that allows a single model to efficiently capture both the predictive performance and uncertainty estimates of an ensemble. For classification, this is achieved by training a Dirichlet distribution over the ensemble members' output distributions via the maximum likelihood criterion. Although theoretically principled, this criterion exhibits poor convergence when applied to large-scale tasks where the number of classes is very high.
In our work, we analyze this effect and show that for the Dirichlet log-likelihood criterion classes with low probability induce larger gradients than high-probability classes. This forces the model to focus on the distribution of the ensemble tail-class probabilities.
We propose a new training objective which minimizes the reverse KL-divergence to a \emph{Proxy-Dirichlet} target derived from the ensemble. This loss resolves the gradient issues of Ensemble Distribution Distillation, as we demonstrate both theoretically and empirically on the ImageNet and WMT17 En-De datasets containing 1000 and 40,000 classes, respectively.
\end{abstract}

\section{Introduction}
\label{sec:introduction}


Ensembles of machine learning models are known to yield improved predictive performance relative to single models~\cite{dietterich2000ensemble}. With the increasing popularity of neural networks, ensemble methods have been rapidly adopted in numerous sub-fields of machine learning~\cite{ashukha2020pitfalls, trust-uncertainty}. More importantly, \cite{deepensemble2017} demonstrated that although single neural networks are often overconfident in their predictions, their ensembles can output reliable uncertainty estimates. Furthermore, ensembles allow \emph{total uncertainty} to be decomposed into \emph{data} and \emph{knowledge uncertainty}\footnote{Data and Knowledge Uncertainty are also known as Aleatoric and Epistemic uncertainty.}. The former is the intrinsic uncertainty due to class overlap and noise inherent in the data, while the latter is the model's uncertainty due to lack of understanding of the test data~\cite{malinin-thesis}. Estimates of \emph{knowledge uncertainty} are often used to detect anomalous and unfamiliar inputs~\cite{batchbald,gal-adversarial, malinin-rkl-2019, malinin-thesis}. Given the increased usage of deep learning for safety-critical applications such as self-driving cars or medical diagnostics, obtaining reliable uncertainty estimates becomes ever more important each year.

However, using ensembles for inference can be computationally prohibitive in certain applications. Obtaining predictions in real time can often be quite involved even for a single model, and the hardware requirements for serving the ensemble of neural networks scale linearly with its size. As a result, over the past several years the area of ensemble distillation has gained increased attention of the research community. Broadly speaking, distillation methods aim to train a single model which can approximate the behavior of the ensemble sufficiently well. 

In the simplest and most frequently used form of distillation \cite{hinton2015distilling}, the student model is trained to capture the average prediction of the ensemble: for example, in case of classification this reduces to KL-divergence between the model and the ensemble mean. While this method allows the student to obtain predictive performance comparable to that of the original ensemble, the information about its distributional properties (in other words, its diversity) is lost in the process. As ensemble-based uncertainty estimation methods often utilize the information about disagreement between its members, such distillation approaches achieve only one of two favorable ensemble properties which we would like to preserve.


Recently, several works have proposed distillation procedures that capture information about both the mean as well as the distribution of ensemble predictions within a single model~\cite{malinin-endd-2019,hydra,mdd,malinin2020regression}. We will broadly refer to this class of distillation approach as \emph{Ensemble Distribution Distillation} (\Endd). Ensemble Distribution Distillation offers a straightforward way to model the ensemble predictions~\cite{malinin-endd-2019,malinin2020regression}. Outputs of each member are viewed as samples from a higher-order Dirichlet or Normal-Wishart distribution, and the student model attempts to learn the parameters of that distribution. Typically, \Endd is done by maximizing the likelihood the ensemble's output distributions under the Dirichlet or Normal-Wishart Prior. While theoretically sound, for large-scale classification tasks with many classes, gradient-based optimization of this criterion is highly problematic, which limits its usefulness in real-life production scenarios.

In this work, we investigate the poor convergence of models trained with Ensemble Distribution Distillation at scale. We analyze the the Dirichlet log-likelihood criterion and show that it leads to high gradient norm values that affect the optimization procedure. Specifically, if a particular ensemble member's output distribution has most probability mass allocated to a few classes, with the remained spread among a long tail of exponentially less-probable classes, then the gradients associated with the tail-classes will be significantly larger than those associated with high-probability classes. As a result, the model focuses on modelling this distribution of probabilities of tail-classes.

To solve this, we propose to transform the empirical distribution of ensemble member predictions into a \emph{Proxy-target} Dirichlet distribution with the same statistics and to use this distribution as the target during distillation. Furthermore, we show that it is crucial to minimize the \emph{reverse} KL-divergence between the model and the Proxy-Dirichlet, as minimization the \emph{forward} KL-divergence exacerbates the optimizations issues. The proposed training procedure allows the model to converge, mitigating the issue of gradient explosion. We demonstrate this by distribution-distilling ensembles of models trained on both the ImageNet classification and WMT17 English-German language translation datasets, where there are 1000 and 40,000 classes, respectively. On both datasets the distribution-distilled models outperforms models trained from scratch and yield uncertainty estimates competitive this those of the original ensemble. 

Thus, our contributions are as follows:
\begin{itemize}
    \item We analyze the issues of Dirichlet distribution likelihood when applied to a large number of classes and confident predictions
    \item We propose several improvements to the {Ensemble Distribution Distillation} framework, each of them arising from the Dirichlet distribution properties in the context of deep learning
    \item We adapt \emph{Ensemble Distribution Distillation} to auto-regressive models and propose {Sequence Ensemble-Distribution Distillation} (SEnD$^2$)
    \item We examine and propose solutions for a range of technical challenges associated with scaling {Ensemble Distribution Distillation} to large output spaces. 
\end{itemize}.
\section{Preliminaries: Ensembles and Distillation}

We view ensembles within a Bayesian framework where the model parameters $\bm{\theta}$ are random variables over which a prior distribution ${\tt p}(\bm{\theta})$ is placed. The posterior distribution ${\tt p}(\bm{\theta}|\mathcal{D})$ is obtained via Bayes' rule:
\begin{empheq}{align}
\begin{split}
  {\tt p}(\bm{\theta}|\mathcal{D}) &= \frac{{\tt p}(\mathcal{D}|\bm{\theta}){\tt p}(\bm{\theta})}{{\tt p}(\mathcal{D})}  \propto {\tt p}(\mathcal{D}|\bm{\theta}){\tt p}(\bm{\theta}) 
\end{split}
\label{eqn:bayesposterior}
\end{empheq}
Consider an ensemble of models $\{{\tt P}(y|\bm{x}^{*}, \bm{\theta}^{(m)})\}_{m=1}^M $ sampled from the posterior:
\begin{empheq}{align}
\begin{split}
\big\{{\tt P}(y| \bm{x}, \bm{\theta}^{(m)} )\big\}_{m=1}^M \rightarrow& \big\{{\tt P}(y| \bm{\pi}^{(m)} )\big\}_{m=1}^M,\quad \bm{\pi}^{(m)} =\ \bm{f}(\bm{x}; \bm{\theta}^{(m)}),\  \bm{\theta}^{(m)}\sim {\tt p}(\bm{\theta}|\mathcal{D})
\end{split}
\end{empheq}
where $\bm{\pi}$ are the parameters of a categorical distribution $[ {\tt P}(y=\omega_1),\cdots, {\tt P}(y=\omega_K)]^{\tt T}$. The predictive distribution, or \emph{predictive posterior}, for a test input $\bm{x}^{*}$ is obtained by taking the expectation with respect to the model posterior:
\begin{empheq}{align}
\begin{split}
    {\tt P}(y| \bm{x}^{*}, \mathcal{D}) = &\ \mathbb{E}_{{\tt p}(\bm{\theta}|\mathcal{D})}\big[{\tt P}(y|\bm{x}^{*}, \bm{\theta})\big]
    \approx \ \frac{1}{M}\sum_{m=1}^M{\tt P}(y|\bm{x}^{*}, \bm{\theta}^{(m)})
\end{split}
\label{eqn:modunc}
\end{empheq}
In practice this is intractable and we approximate via Monte-Carlo sampling. Given the ensemble, the entropy of the predictive posterior is a measure of \emph{total uncertainty}. \emph{Knowledge uncertainty} can be assessed via measures of the spread, or `disagreement', of the ensemble such as \emph{mutual information}:
\begin{empheq}{align}
\begin{split}
\underbrace{\mathcal{I}[y,\bm{\theta}| \bm{x}^{*},\mathcal{D}]}_{\text{Knowledge Uncertainty}} = &\ \underbrace{\mathcal{H}\big[\mathbb{E}_{{\tt p}(\bm{\theta}|\mathcal{D})}[{\tt P}(y|\bm{x}^{*}, \bm{\theta})]\big]}_{\text{Total Uncertainty}} - \underbrace{\mathbb{E}_{{\tt p}(\bm{\theta}|\mathcal{D})}\big[\mathcal{H}[{\tt P}(y|\bm{x}^{*},\bm{\theta})]\big]}_{\text{Expected Data Uncertainty}} 
\end{split}
\label{eqn:mibayes}
\end{empheq}

While ensembles yield improved predictive performance and theoretically interpretable uncertainty estimates, they are expensive during training, and especially so during inference. Thus, it is common to \emph{distill} an ensemble into a single model. Typically, this is done by minimizing the KL-divergence to the predictive posterior of the ensemble:
\begin{empheq}{align}
\begin{split}
\mathcal{L}^{\text{EnD}}(\bm{\phi},\mathcal{D}_{\tt ens}) =&  \mathbb{E}_{{\tt \hat p}(\bm{x})}\Big[{\tt KL}\big[{\tt P}(y| \bm{x}, \mathcal{D})\ ||\ {\tt P}(y| \bm{x};\bm{\phi})\big] \Big]
\end{split}
\end{empheq}
This approach has been thoroughly investigated for a range of tasks, such as image classifcation, machine translation, etc..
While distillation allows a single model to capture the predictive quality and estimates of \emph{total uncertainty} of the ensemble at low computational and memory cost, information about the diversity of the ensemble is lost. Consequently, it is no longer possible to obtain estimates of \emph{knowledge uncertainty} which is particularly useful for anomaly detection~\cite{malinin-thesis, malinin-endd-2019}.

\cite{malinin-endd-2019} recently proposed a class of distillation techniques called \emph{Ensemble Distribution Distillation} (\Endd), where the goal is to capture both the mean and the diversity of an ensemble within a single model. The proposed solution to \Endd was to distill an ensemble into a Prior Network model which parameterizes the Dirichlet distribution as follows:
\begin{empheq}{align}
\begin{split}
{\tt p}(\bm{\pi} | \bm{x};\bm{\hat \phi}) =& {\tt Dir}(\bm{\pi} | \bm{\hat \alpha}), \bm{\hat \alpha} = e^{\bm{z}}, \bm{z}= \bm{f}(\bm{x};\bm{\hat \phi}),\
\hat \alpha_c > 0,\ \hat \alpha_0 = \sum_{c=1}^K \hat \alpha_c
\end{split}
\label{eqn:DPN1}
\end{empheq}


Distribution distillation is then accomplished as follows. Firstly, a \emph{transfer dataset} $\mathcal{D}_{\tt ens}= \{\bm{x}^{(i)}, \bm{\pi}^{(i,1:M)} \}_{i=1}^N \sim {\tt \hat p}(\bm{x},\bm{\pi})$ is composed of the inputs $\bm{x}_i$ from the original training set $\mathcal{D}=\{\bm{x}^{(i)},y^{(i)}\}_{i=1}^N$ and the categorical distributions $\{\bm{\pi}^{(i,1:M)}\}_{i=1}^N$ derived from the ensemble for each input. Secondly, given this transfer set, the model ${\tt p}(\bm{\pi} | \bm{x};\bm{\phi})$ is trained by minimizing the negative log-likelihood of each categorical distribution $\bm{\pi}^{(im)}$:
\begin{empheq}{align}
\begin{split}
\mathcal{L}^{\text{EnD}^2}(\bm{\phi},\mathcal{D}_{\tt ens}) =&\  -\mathbb{E}_{{\tt \hat p}(\bm{x})}\big[\mathbb{E}_{{\tt \hat p}(\bm{\pi}|\bm{x})}[\ln{\tt p}(\bm{\pi} | \bm{x};\bm{\phi}) ] \big] 
\end{split}
\label{eqn:endd-loss1}
\end{empheq}
Given a distribution-distilled Prior Network, the predictive distribution is given by the expected categorical distribution $\bm{\hat \pi}$ under the Dirichlet prior:
\begin{empheq}{align}
\begin{split} 
{\tt P}(y = \omega_c| \bm{x}^{*};\bm{\hat \phi}) = &\ \mathbb{E}_{{\tt p}(\bm{\pi} | \bm{x}^{*};\bm{\hat \phi})}[{\tt P}(y = \omega_c | \bm{\pi})]=\ \hat \pi_c\ = \frac{\hat \alpha_c}{\sum_{k=1}^K \hat \alpha_k} =\ \frac{ e^{\hat z_c}}{\sum_{k=1}^K e^{\hat z_k}}
\end{split}\label{eqn:dirposterior}
\end{empheq}
Measures of \emph{total} and \emph{knowledge uncertainty} are obtained by considering the mutual information between the prediction $y$ and the parameters of $\bm{\pi}$ of the categorical: 
\begin{empheq}{align}
\begin{split}
   \underbrace{\mathcal{I}[y,{\tt \bm{\pi}} |\bm{x}^{*};\bm{\hat \phi}]}_{\text{Knowledge Uncertainty}}=&\ \underbrace{\mathcal{H}\big[\mathbb{E}_{{\tt p}({\tt \bm{\pi}}|\bm{x}^{*};\bm{\hat \phi})}[{\tt P}(y|{\tt \bm{\pi}})]\big]}_{\text{Total Uncertainty}} - \underbrace{\mathbb{E}_{{\tt p}({\tt \bm{\pi}}|\bm{x}^{*};\bm{\hat \phi})}\big[\mathcal{H}[{\tt P}(y|{\tt \bm{\pi}})]\big]}_{\text{Expected Data Uncertainty}} 
\end{split}
 \label{eqn:mipn}
\end{empheq}

It is important to highlight that \Endd can also be accomplished by distilling an ensemble into a mixture model which yields a separate softmax for each ensemble member~\cite{hydra,mdd}. The principle downside of this approach is that it requires more parameters, and attempts to model the ensemble in excessive detail, which requires more flexible and powerful models. As a result, for good performance, it necessary to split the model into multiple heads at an earlier stage, which significantly increases computational and memory complexity. In contrast, \Endd via Prior Networks has a fixed computational and memory cost of one model regardless of the size of the original ensemble.

\begin{figure}[ht]
    \centering
    \subfigure[Initialization]{\includegraphics[width=0.32\textwidth]{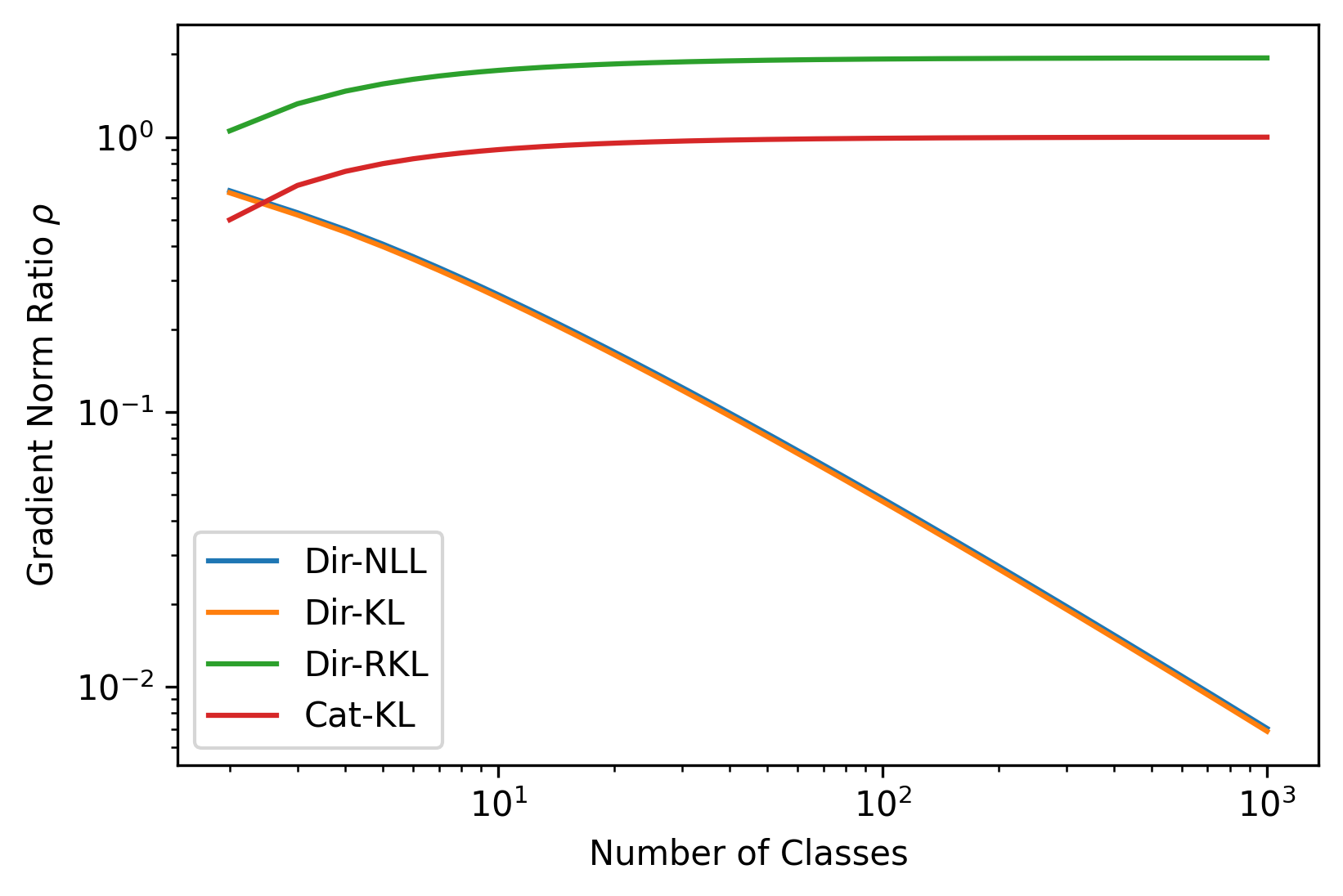}}
    \subfigure[Near Convergence]{\includegraphics[width=0.32\textwidth]{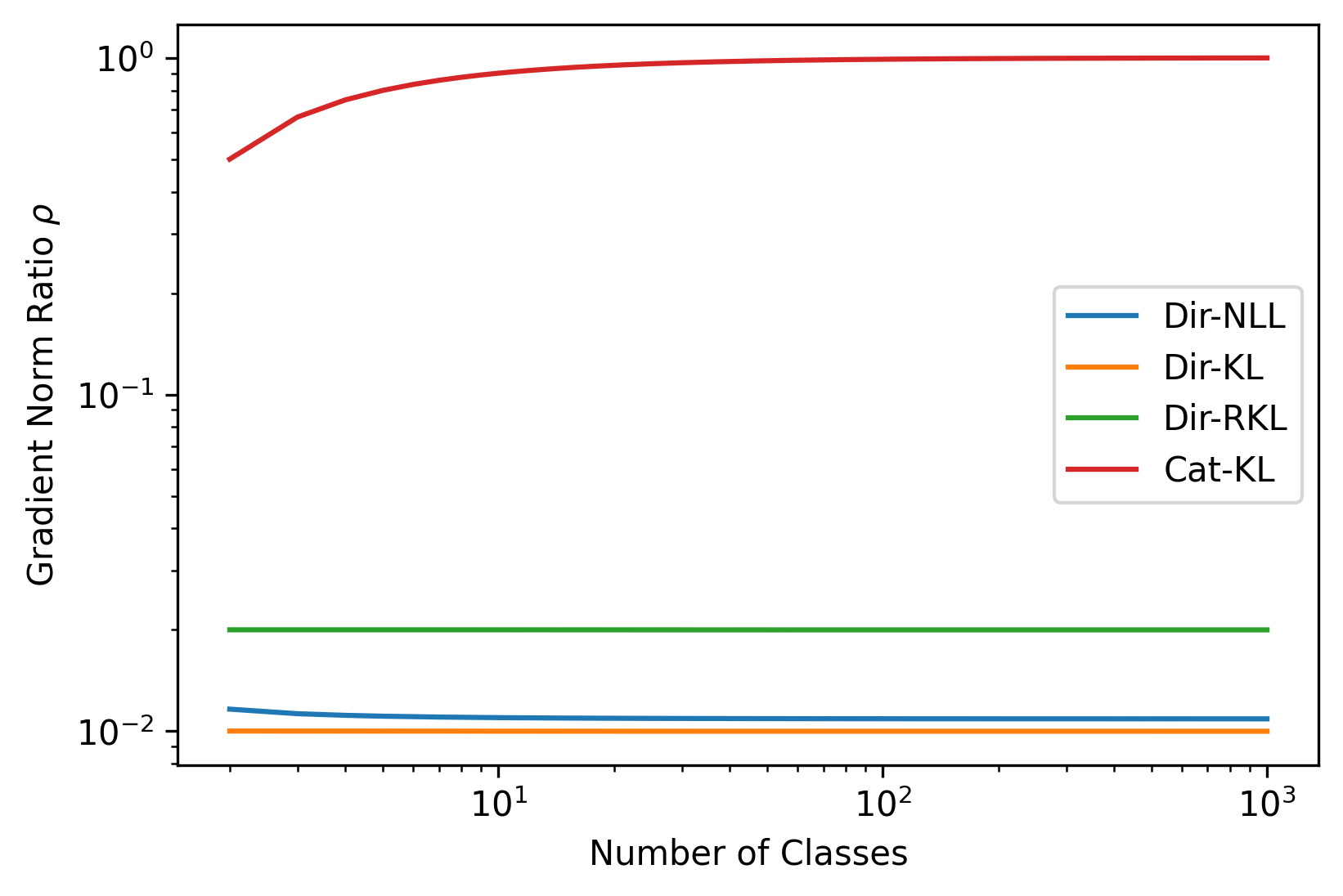}}
    \subfigure[Misclassification]{\includegraphics[width=0.32\textwidth]{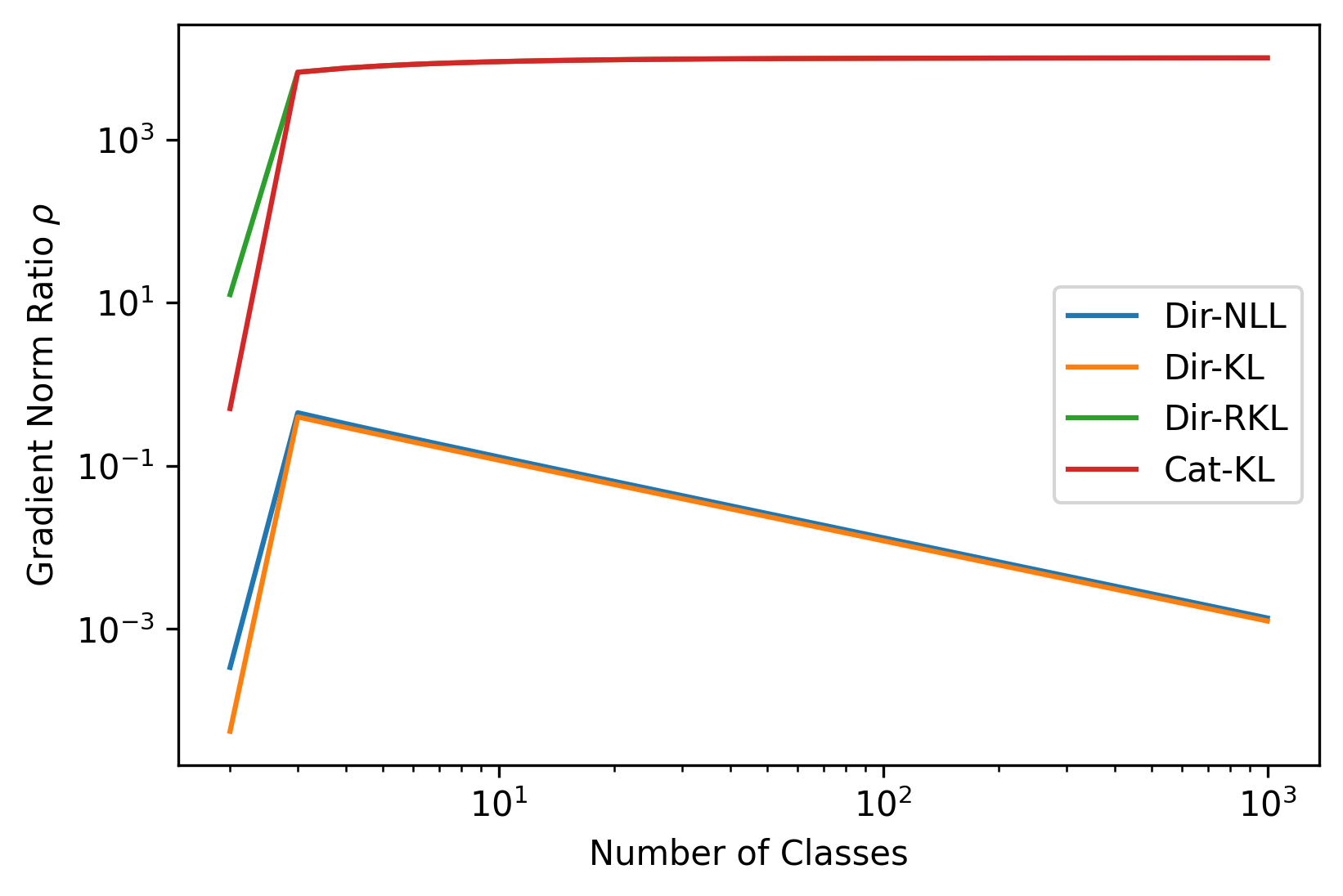}}
    \caption{Gradient Ratio}
    \label{fig:grad_ratio}
\end{figure}

\section{Theoretical Analysis and Alternative Loss functions}

In the previous section we described how \Endd can be done by maximising the log-likelihood of the ensemble's output distributions under a conditional Dirichlet Prior. However, we empirically observed significant convergence issues when applying this approach to tasks with large numbers of classes. Thus, in this section we examine the gradients of the Dirichlet NLL loss and propose an alternate training approach which overcomes them.

\textbf{First-Order Analysis}

The setup which will consider in our analysis is the following. First, we have a Prior Network model which is initialized such that it always returns a uniform Dirichlet distribution ($\bm{\alpha} = \bm{1}$), while the target distribution whose probability is being maximized is a sparse K-length vector of probabilities:
\begin{empheq}{align*}
    \bm{\pi}_{tgt} = \big[1-\epsilon, \epsilon/(K-1), \epsilon/(K-1), \cdots \big]^{\tt T},\quad \epsilon = \text{1e-4}
\end{empheq}
Second, we have a Prior Network which is \emph{near convergence} with the following output distribution:
\begin{empheq}{align*}
    \bm{\alpha}_{cnv} =&\ \bm{\pi}_{cnv} \cdot \alpha_0,\ \alpha_0 = 90K,\quad 
    \bm{\pi}_{cnv} =\ \big[1-5\epsilon, \frac{5\epsilon}{K-1}, \frac{5\epsilon}{K-1}, \cdots \big]^{\tt T}
\end{empheq}
Finally, we have a Prior Network which has made a strong mistake, which represents a situation which could occur somewhere in the middle on training, far from convergence:
\begin{empheq}{align*}
    \bm{\alpha}_{msc} =&\ \bm{\pi}_{msc} \cdot \alpha_0,\ \alpha_0 = 90K,\quad 
    \bm{\pi}_{msc} =\ \big[\frac{5\epsilon}{K-1}, \frac{5\epsilon}{K-1}, \cdots, 1-5\epsilon \big]^{\tt T}
\end{empheq}

First, lets consider the standard cross-entropy loss between a predicted and target discrete distribution and it's gradient with respect to the logit $z_k$:
\begin{empheq}{align}
    \mathcal{L}^{\text{CE}} =&\ -\sum_{k=1}^K \hat \pi_k \ln\big(\frac{\alpha_k}{\alpha_0}\big),\quad 
    \frac{\partial\mathcal{L}^{\text{CE}}}{\partial z_k} =\ \frac{\alpha_k}{\alpha_0} - \hat \pi_k 
\end{empheq}

Second, consider the NLL loss of a Dirichlet distribution and its gradient with respect to logit $z_k$:
\begin{empheq}{align}
   \mathcal{L} \small{=} \sum_{k=1}^K\Gamma(\alpha_k) \small{-}(\alpha_k \small{-} 1)\sum_{m=1}^M\frac{\ln\pi_k^{(m)}}{M} \small{-} \Gamma(\alpha_0), \   \frac{\partial\mathcal{L}}{\partial z_k} \small{=} \big(\psi(\alpha_k) \small{-} \psi(\alpha_0)  \small{-}\sum_{m=1}^M\frac{\ln\pi_k^{(m)}}{M}\big) \cdot \alpha_k
\end{empheq}

Finally, consider the dimensionality normalized ratio of the gradient with respect to the logit 1 and logit 2, which represents the relative contribution of the gradients with respect to the class we are interested in modelling to the long tail. 
\begin{empheq}{align}
\begin{split}
        \rho = \frac{1}{K} \Big| \frac{\partial\mathcal{L}}{\partial z_1} \Big| \Big/ \Big|\frac{\partial\mathcal{L}}{\partial z_2}\Big|
\end{split}
\end{empheq}
Figure~\ref{fig:grad_ratio} shows that, at initialization, as the number of classes is increased the standard cross-entropy loss primarily focuses on the high probability class and ignores the long tail. In contrast, the Dirichlet NLL loss displays a diminishing contribution. This means that the loss will focus on modelling the probability distribution of the high-probability classes only after it \emph{perfectly} models the long tail. As the loss is also very sensitive, it means that on complex tasks the model is perpetually stuck modelling the probabilities of tail classes. Note that even near convergence, the ratio $\rho$ is far smaller for the NLL criterion than for discrete cross-entropy. Finally, if a significant error is made on the training data, $\rho$ becomes very large for cross-entropy, and increasingly small for Dirichlet NLL as the number of classes increases. This analysis shows that a desirable property of the loss which ensures good convergence is that the ratio $\rho$ is high and either constant or increasing as the number of classes grows, otherwise the model focuses on modelling the distribution of tail-class probabilities across the ensemble.

An additional issue to consider is that the NLL noise is also noisy, as for each input $\bm{x}$ we only have a few discrete distributions - it may be necessary to use far more samples to get a good estimate of the ensemble's distribution. Furthermore, this distribution may be poorly matched to the Dirichlet, which introduces additional issues. Thus, a natural solution to consider would be to introduce a \emph{Proxy Dirichlet Distribution} to which we can minimize either the \emph{KL-divergence} or \emph{reverse KL divergence}. We leave discussion of the details of the Proxy Dirichlet until later and only consider the gradients which arise from minimizing either loss.  

For this analysis we consider a target Dirichlet distribution with parameters $\bm{\beta} = \bm{\pi}_{tgt}*\beta_0$ where $\beta_0 = 100K$. The explicit forms of the KL-divergence between two Dirichlet distributions, as well the gradient of the forward and reverse KL-divergence are provided below:
\begin{empheq}{align}
\begin{split}
       & \mathcal{L}^{\text{KL}} =\ \ \sum_{k=1}^K\Gamma(\alpha_k) - \sum_{k=1}^K\Gamma(\beta_k) + \Gamma(\beta_0) - \Gamma(\alpha_0) + \sum_{k=1}^K(\beta_k - \alpha_k)\Big(\psi(\beta_k)-\psi(\beta_0)\Big)
\end{split} \\
\begin{split}
       & \mathcal{L}^{\text{RKL}} =\ \ \sum_{k=1}^K\Gamma(\beta_k) - \sum_{k=1}^K\Gamma(\alpha_k) + \Gamma(\alpha_0) - \Gamma(\beta_0) + \sum_{k=1}^K(\alpha_k - \beta_k)\Big(\psi(\alpha_k)-\psi(\alpha_0)\Big)
\end{split} \\
\begin{split}
        &\frac{\partial\mathcal{L}^{\text{KL}}}{\partial z_k} =\ \big(\psi(\alpha_k) - \psi(\alpha_0) - \psi(\beta_k) + \psi(\beta_0)\big) \cdot \alpha_k
\end{split} \\
\begin{split}
        &\frac{\partial\mathcal{L}^{\text{\tiny RKL}}}{\partial z_k} =\ \big((\alpha_k - \beta_k)\psi'(\alpha_k) - (\alpha_0 - \beta_0)\psi'(\alpha_0)\big) \cdot \alpha_k
\end{split}
\end{empheq}

Figure~\ref{fig:grad_ratio} additionally displays the ratio $\rho$ for both the forward and reverse KL-divergence losses. The forward KL-divergence displays the same issues as the NLL loss and $\rho$ continues to decrease as the number of classes in increased. This is unsurprising, as the NLL is equivalent to the KL-divergence in the limit. However, the \emph{reverse KL-divergence} displays the desirable properly that $\rho$ grows and stabilizes as the number of classes is increased. This suggests that if we were to minimize the \emph{reverse KL-divergence} to an appropriately chosen \emph{Proxy-Target Dirichlet distribution}, then we would be able to avoid convergence issues. 



\textbf{Proxy-Dirichlet distribution}

\begin{figure*}[ht]
    \centering
    \subfigure[Naive \Endd]{\includegraphics[scale=0.067]{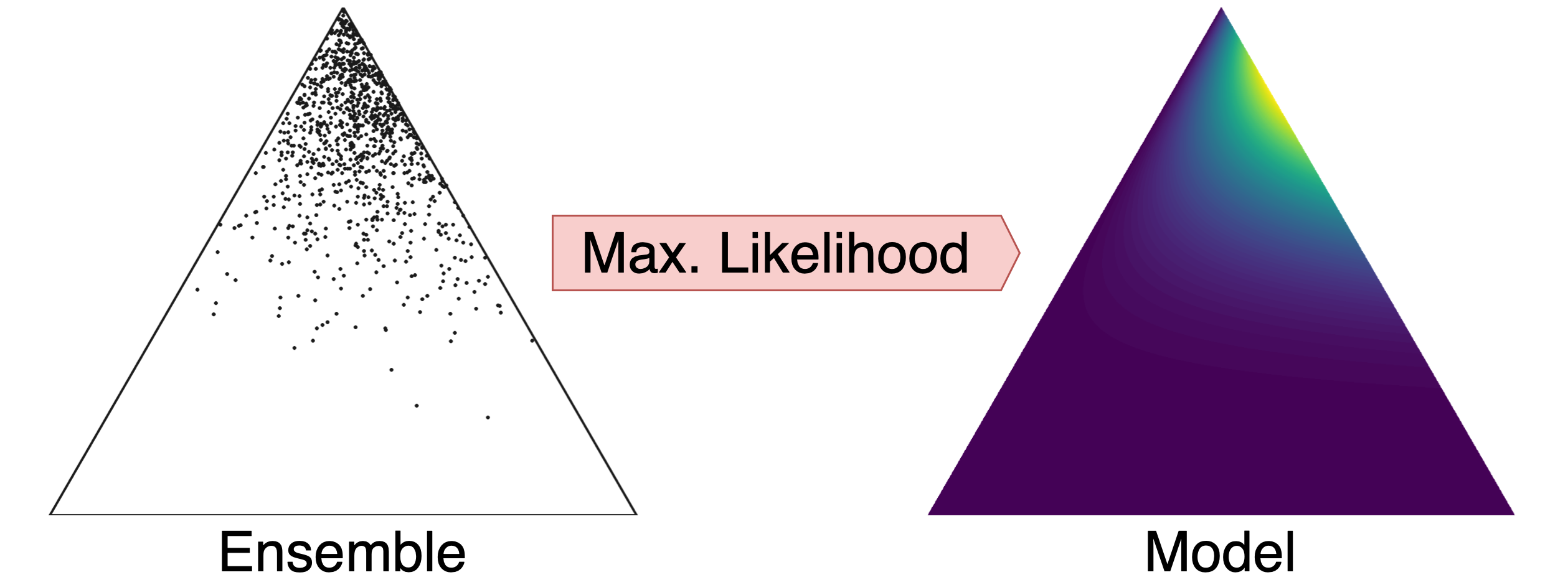}}
    \subfigure[Proxy \Endd]{\includegraphics[scale=0.067]{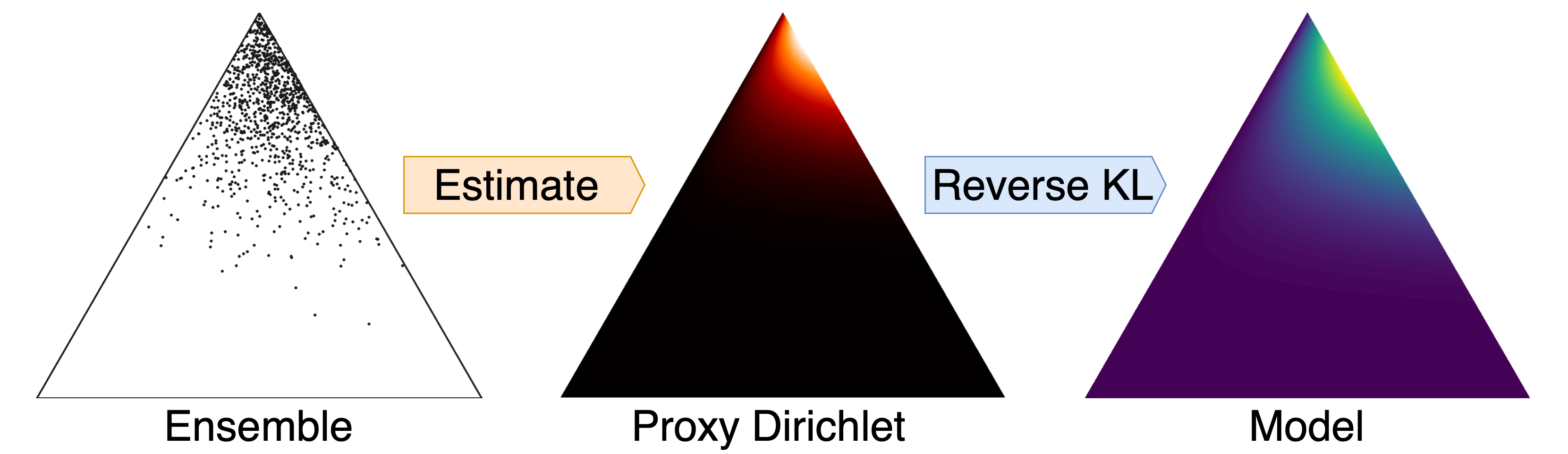}}
    \caption{Schematic of Distillation Approaches}
    \label{fig:distillation overview}
\end{figure*}

It is important to remember that the ensemble may be poorly modelled via a Dirichlet distribution, so it is necessary to ask which properties of the ensemble we are actually interested in capturing. Clearly, we would like to capture the mean of the ensemble, as that typically has better predictive accuracy and calibration. Additionally, we would like to capture \emph{bulk-diversity properties} of the ensemble, such that the measures of divergence derived from the Proxy Dirichlet are similar to those of the original ensemble and therefore provide a similar rank-ordering of data. At the same time, we are \emph{not} interested modelling properties like multi-modality and skew. 

Clearly, obtaining the mean of the ensemble is trivial. Obtaining an estimate of the precision $\beta_0$ is more challenging. One approach based on Sterling's approximation is described in~\cite{minka2000estimating} and proposes the following estimate:
\begin{empheq}{align}
\begin{split}
        \hat \pi_k (\bm{x})=&\ \frac{1}{M}\sum_{m=1}^M {\tt P}(y=\omega_k|\bm{x}, \bm{\theta}^{(m)}) \\
        \tilde \beta_0(\bm{x}) =& \frac{K-1}{2 \sum_{k=1}^K\hat \pi_k (\ln \hat \pi_k - \frac{1}{M}\sum_{m=1}^M\ln \pi_k^{(m)})},\ \bm{\beta}_k (\bm{x}) = \ \hat \pi_k(\bm{x}) \cdot \tilde \beta_0(\bm{x}) + 1
\end{split}
\end{empheq}

\begin{figure}[ht]
    \centering
    \subfigure[Initialization]{\includegraphics[scale=0.49]{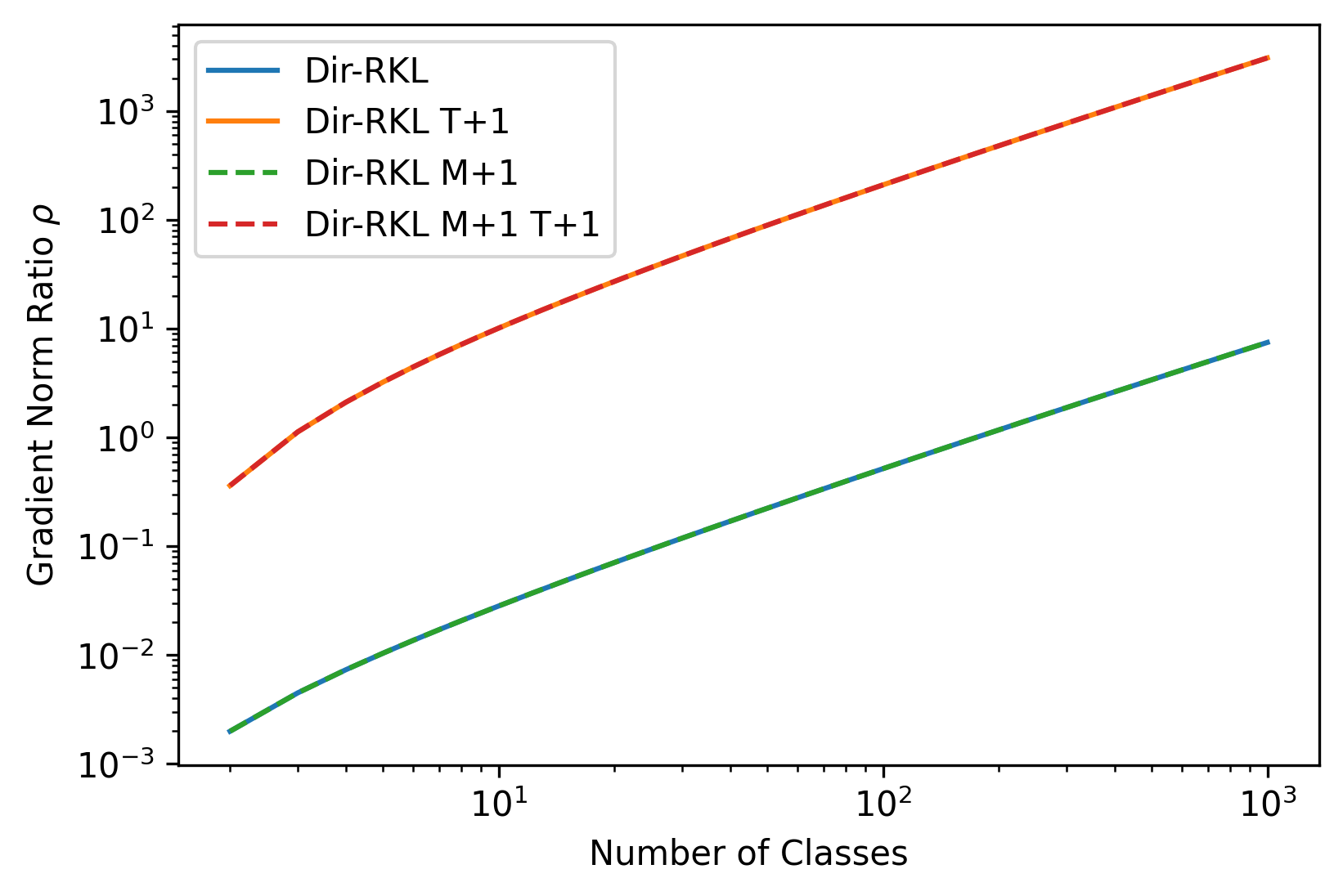}}
    \subfigure[Near Convergence]{\includegraphics[scale=0.49]{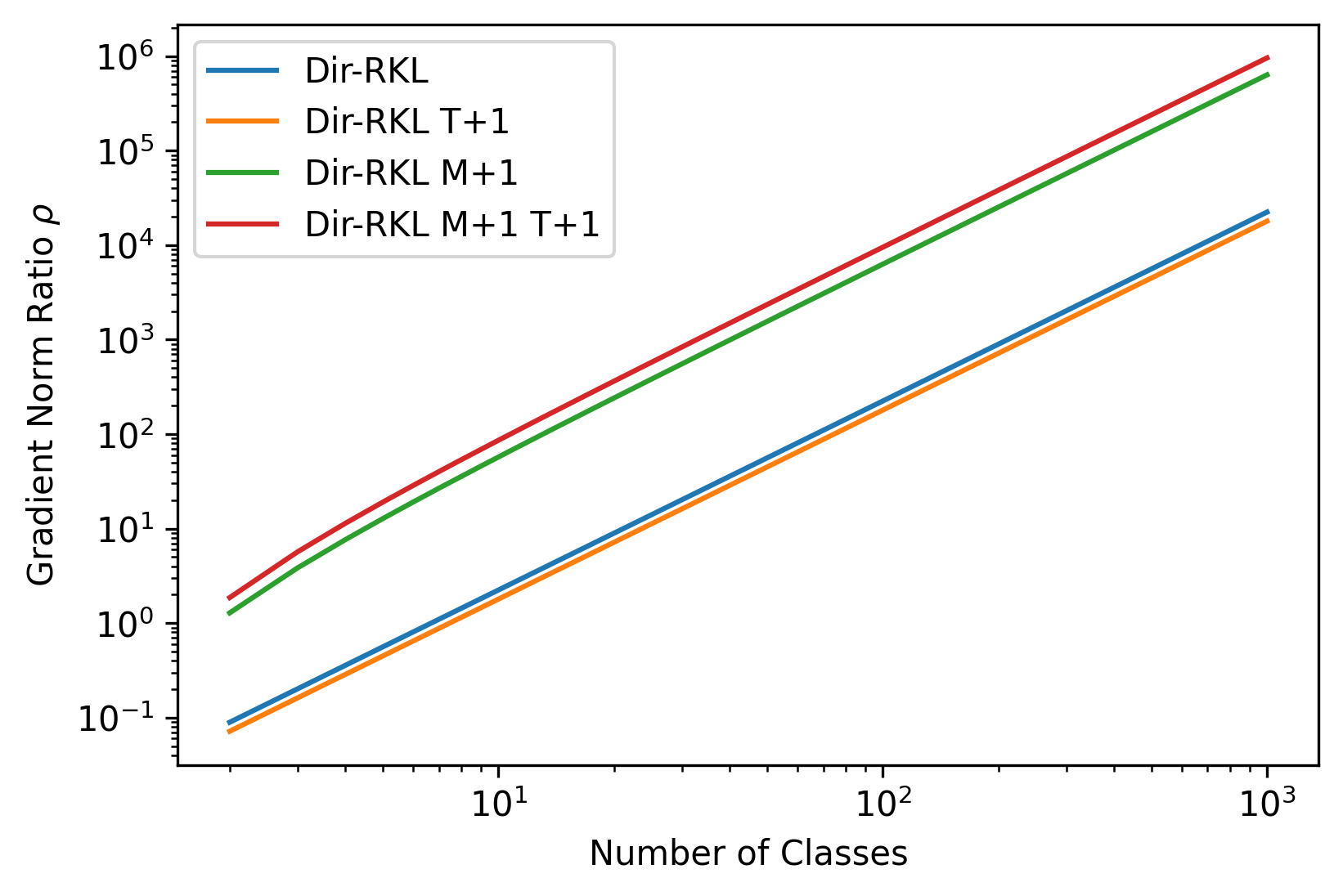}}
    \caption{Gradient Ratio}
    \label{fig:grad_ratio_smooth}
\end{figure}

We found that it is important to also add 1 to all the target concentration parameters. Figure~\ref{fig:grad_ratio_smooth} shows that for the reverse KL loss, adding 1 to \emph{both} the target Proxy-Dirichlet as well as \emph{the model} yields an improved ratio $\rho$ both at initialization and near convergence. Heuristically, it seems to make the loss more linear and stable by preventing the digamma and trigamma functions $\psi$ and $\psi'$ in the reverse-KL loss from dropping into the highly non-linear regime when $\alpha_k < 1$ and $\beta_k < 1$.

Note, while the solution may seem to similar to work done in \cite{malinin-rkl-2019}, the fundamental underlying reason for using this loss is altogether different. Here, the issue is due to large gradients from low-probability tail classes, while in~\cite{malinin-rkl-2019} the reverse KL loss is used to avoid inducing a multi-modal target Dirichlet distribution in expectation. 

\begin{empheq}{align}
\begin{split}
{\tt KL}[{\tt p}(\bm{\pi}|\bm{x},\bm{\theta}) \| {\tt p}(\bm{\pi}|\bm{\hat \beta})] =&\ \underbrace{ \beta_0\cdot\mathbb{E}_{{\tt p}(\bm{\pi}|\bm{x},\bm{\theta})}\big[-\sum_{k=1}^K\hat \pi_k\ln \pi_k\big]}_{\text{Reconstruction term}} + \underbrace{{\tt KL}[{\tt p}(\bm{\pi}|\bm{x},\bm{\theta}) \| {\tt p}(\bm{\pi}|\bm{1})]}_{\text{Prior}} +Z
\end{split}
\end{empheq}


\section{Experiments}
\label{sec:experiments}

In this section, we evaluate \Endd via minimization of Reverse KL-divergence between the model and a Proxy Dirichlet target. We apply distribution distillation to ensembles of convolutional networks trained on the ImageNet dataset and to ensemble of Transformer models trained on WMT'17 En-De. Our goal here is to demonstrate that given an ensemble, we can successfully distribution-distill it into a single model. Note that we do not provide results for \Endd accomplished by optimizing Dirichlet NLL or forward KL-divergence, because we could not get them to even begin to converge on the tasks considered here. 

\subsection{Setup}
\label{sec:experiments_setup}
We consider two large-scale tasks involving classification: 1000-class image classification and sequence-to-sequence modeling of natural language. For each task, we first train the ensemble of regular models and then distill it with \Endd. For comparison, we also report the average single-model performance along with the following baselines:

\begin{itemize}
\item \textbf{Ensemble} refers to the performance of an ensemble of independently trained models, which was previously shown to yield high quality uncertainty estimates~\cite{deepensemble2017} and to outperform more sophisticated methods using only a few models~\cite{ashukha2020pitfalls}.
\item \textbf{Ensemble Distillation} (EnD) is a common approach to model and ensemble distillation, first proposed in~\cite{hinton2015distilling}. It involves training the student model with the soft target distribution of averaged ensemble predictions. Notably, we do not add the cross-entropy loss for ground truth labels, because we focus on the comparison of distillation objectives and not only classification performance.
\end{itemize}

We do not use Hydra~\cite{hydra} or similar multi-head approaches for distilling each separate ensemble member, because with a high number of models in the ensemble and even 1000 classes the computation overhead is no longer negligible. In all experiments with \Endd, we add 1 both to the predicted parameters of the Dirichlet distribution and the Dirichlet proxy parameters.

Both for error rejection and out-of-distribution detection, we use several information-theoretic measures uncertainty; in particular, we use entropy of the expected predictive distribution (EoE) for total uncertainty and Reverse Mutual Information (RMI) for knowledge uncertainty throughout this section.
Derivations of these measures both for \Endd and ensembles are available in~\cite{malinin-thesis} and~\cite{malinin-structured-2020}.
For Single and EnD single-model baselines, we use
entropy of the output distribution as the only valid uncertainty estimate.

\subsection{Large-scale image classification}
\label{experiments:imagenet}

For the first experiment, we run distillation of the ensemble that contains 10 ResNet-50~\cite{resnet} models trained on the ImageNet~\cite{imagenet} image classification dataset. We use the standard training setup outlined in~\cite{touvron2019FixRes}; specifically, we train for 90 epochs using stochastic gradient descent with momentum of 0.9 and a learning rate of $0.1\times B/256$ (first proposed in~\cite{goyal2018accurate}), where B is the per-device batch size multiplied by the number of GPUs. 
In our experiments, we use a single-GPU batch size of 256 and 8 NVIDIA V100 GPUs. The learning rate is divided by 10 every 30 epochs. For data augmentations, we use a standard combination of random resized crops and horizontal flips implemented in the Albumentations library~\cite{albumentations}.
In all experiments, we found it beneficial to initialize the last batch normalization $\gamma$ in each residual branch to zero, which agrees with previous results~\cite{goyal2018accurate, zhang2018residual, rezero}.

For a thorough evaluation of all methods, we use several different characteristics of performance. First, we measure the in-domain classification accuracy on the original ImageNet validation subset~\cite{imagenet}, which is commonly used for comparison of image classification models. Second, we compare the robustness of all approaches to different domain shifts, also measured by accuracy on datasets corresponding to these shifts. In particular, we use adversarial examples from ImageNet-A~\cite{hendrycks2021nae}, corrupted and perturbed versions of original ImageNet validation data from ImageNet-C~\cite{hendrycks2019robustness}, and artistic renditions from ImageNet-R~\cite{hendrycks2020many}. Next, these domain shift and the original validation dataset are used to compare calibration of models with Expected Calibration Error (ECE).
Finally, we measure the out-of-distribution detection error in terms of Receiver Operating Characteristic area under curve (ROC AUC) on the domain shift datasets together with ImageNet-O~\cite{hendrycks2021nae}.



\begin{figure}
    \centering
    \includegraphics[width=\textwidth]{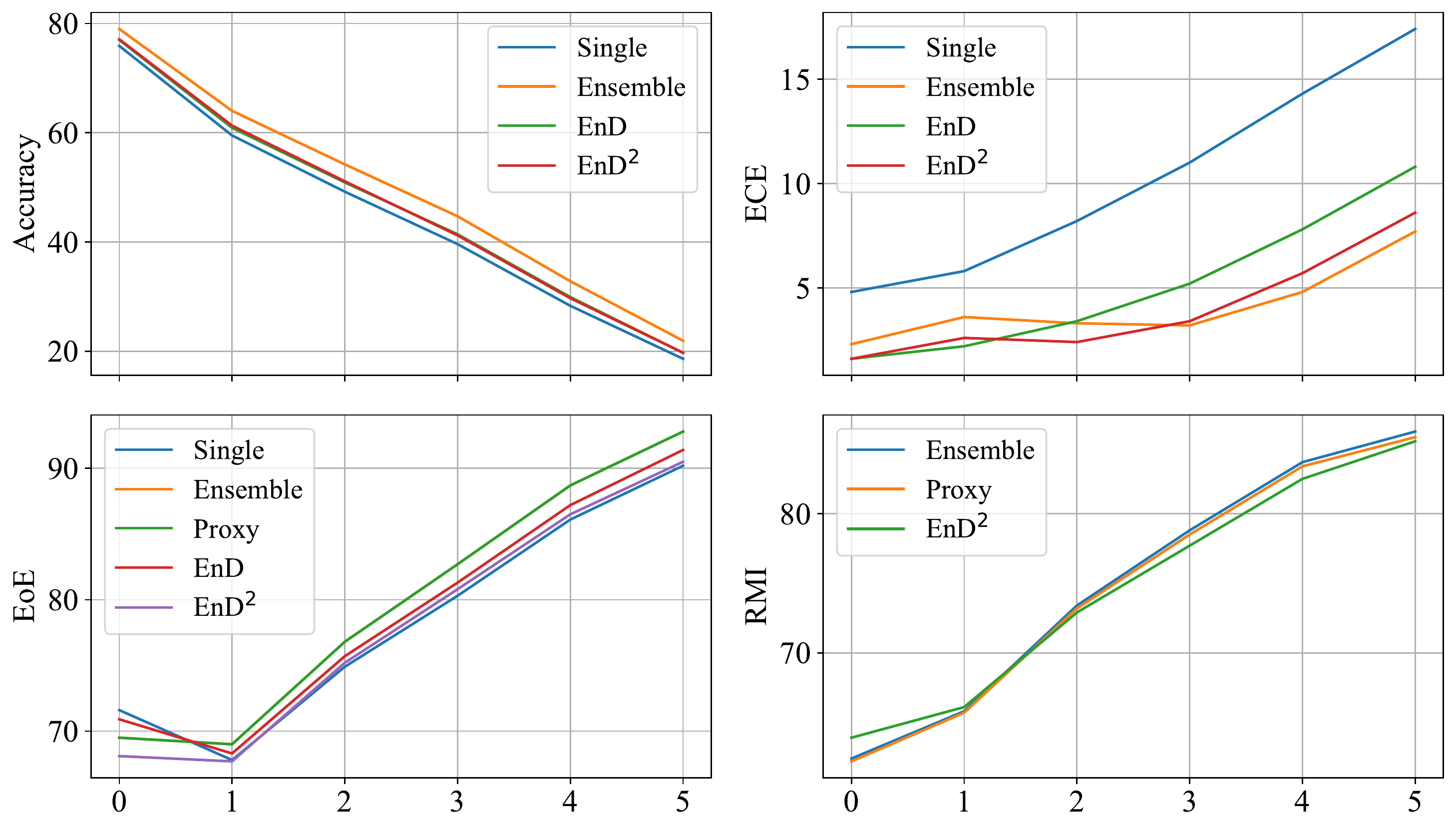}
    \caption{Performance of image classification models depending on the level of ImageNet-C corruption.  No corruption corresponds to the original ImageNet validation data.}
    \label{fig:imagenet_breakdown}
\end{figure}

We report the results for all metrics in Tables~\ref{tab:imagenet_pred} and~\ref{tab:imagenet_ood} for prediction quality and out-of-distribution detection respectively.
Here, the metrics on ImageNet-C are averaged over all degrees of corruption; in Figure~\ref{fig:imagenet_breakdown}, we provide the detailed results of evaluation on each degree separately.
For out-of-distribution detection, we also provide the results of the Dirichlet Proxy to verify that this approximation of the ensemble predictive distribution does not significantly affect its performance.

Table~\ref{tab:imagenet_pred} shows that \Endd is capable of accurate emulation of the ensemble in terms of classification performance: in terms of accuracy, the method displays results on par or slightly better than regular distillation while also having smaller calibration errors. Also, in Table~\ref{tab:imagenet_ood}, it can be seen that for most datasets (except the hardest ImageNet-O) Proxy-Dirichlet distillation can closely match the out-of-distribution performance of the ensemble. As expected, both distillation methods outperform training a single model from scratch while having the same computational complexity.

Furthermore, Figure~\ref{fig:imagenet_breakdown} shows that as the domain shift increases, all models suffer from a drop in accuracy and calibration quality; notably, EnD and \Endd have the same calibration performance on original data, but Dirichlet network distillation has lower calibration errors for the highest degrees of corruption. Unsurprisingly, the further the data is from the original training images, the better the models are at out-of-distribution detection.

\begin{table}
\centering
\small
\caption{Prediction quality results for image classification.}
\label{tab:imagenet_pred}
\begin{tabular}{lcccccccc}
\toprule
{} & \multicolumn{2}{c}{ImageNet-val} & \multicolumn{2}{c}{ImageNet-A} & \multicolumn{2}{c}{ImageNet-C} & \multicolumn{2}{c}{ImageNet-R} \\
{} &          Acc &      ECE &        Acc &       ECE &        Acc &       ECE &        Acc &       ECE \\
\midrule
Single   &     75.9±0.1 &  4.8±0.1 &    4.4±0.2 &  51.1±0.3 &   39.1±0.7 &  11.3±0.7 &   35.0±0.2 &  21.3±0.4 \\
Ensemble &         79.0 &      2.3 &        3.9 &      42.0 &       43.5 &       4.5 &       38.8 &       9.8 \\
EnD      &         77.0 &      1.6 &        3.8 &      46.6 &       40.6 &       5.9 &       36.9 &      16.1 \\
\Endd    &         77.1 &      1.6 &        3.9 &      42.8 &       40.6 &       4.5 &       37.0 &      11.8 \\
\bottomrule
\end{tabular}
\end{table}

\begin{table}
\centering
\small
\caption{Out-of-distribution detection results for image classification.}
\label{tab:imagenet_ood}
\begin{tabular}{lcccccccc}
\toprule
{} & \multicolumn{2}{c}{ImageNet-O} & \multicolumn{2}{c}{ImageNet-A} & \multicolumn{2}{c}{ImageNet-C} & \multicolumn{2}{c}{ImageNet-R} \\
{} &        EoE &   RMI &        EoE &   RMI &        EoE &   RMI &        EoE &   RMI \\
\midrule
Single   &   50.7±0.3 &     - &   85.8±0.1 &     - &   79.9±0.4 &     - &   83.0±0.2 &     - \\
Ensemble &       54.6 &  62.7 &       88.8 &  86.7 &       82.0 &  77.5 &       86.1 &  84.1 \\
Proxy    &       54.6 &  62.9 &       88.8 &  86.5 &       82.0 &  77.3 &       86.1 &  84.0 \\
EnD      &       48.4 &     - &       87.2 &     - &       80.8 &     - &       83.9 &     - \\
\Endd    &       52.0 &  53.2 &       86.8 &  84.6 &       80.1 &  76.9 &       83.7 &  81.4 \\
\bottomrule
\end{tabular}
\end{table}

\subsection{Machine translation}
\label{experiments:nmt}
For this experiment, we train standard Transformer-big~\cite{vaswani2017attention} models on the WMT'17 English-German machine translation dataset with the vocabulary of 40,000 Byte-Pair Encoding tokens~\cite{sennrich-etal-2016-neural}. Each of the 10 ensemble members is trained with the setup described in~\cite{ott2018scaling}: in particular, we train them for 193,000 steps with Adam~\cite{adam} on 8 NVIDIA V100 GPUs with a batch size of 4096 tokens per GPU. We train all distillation models for 20,000 steps with the increased batch size of 32K tokens. Because our approach requires fitting all 10 ensemble members in GPU memory, we reduce the immediate batch size for each step to 1024, but compensate for it with gradient accumulation over 32 steps. For output generation and estimation of uncertainty measures (where applicable), we use beam search with beam size 5.

To compare the approaches in terms of translation quality, we use the BLEU score~\cite{papineni2002bleu} computed with SacreBLEU~\cite{sacrebleu} and sequence-level Prediction Rejection Ratio~\cite{malinin-thesis} on the newstest14 English-German test set. For out-of-distribution detection, we also compute ROC AUC and use several datasets with different characteristics and degrees of domain shift: sentences with permuted tokens in the input, LibriSpeech~\cite{librispeech} test-clean speech transcriptions, and source sentences from newstest14 in German and French languages respectively. We average the results of both distillation methods over 5 random seeds and provide standard deviations of all metrics.

\begin{table}
\centering
\small
\caption{Prediction quality results for machine translation.}
\label{tab:wmt_pred}
\begin{tabular}{lccc}
\toprule
{} &      BLEU &       EoE &       RMI \\
\midrule
Single   &  28.8±0.1 &  36.0±1.3 &  - \\
Ensemble &      30.1 &      30.2 &      26.0 \\
EnD      &  29.4±0.1 &  35.6±0.4 &  - \\
\Endd    &  29.5±0.1 &  35.9±0.8 &  35.8±0.5 \\
\bottomrule
\end{tabular}
\end{table}

\begin{table}
\centering
\small
\caption{Out-of-distribution detection results for machine translation.}
\label{tab:wmt_ood}
\begin{tabular}{lcccccccc}
\toprule
{} & \multicolumn{2}{c}{Permuted} & \multicolumn{2}{c}{Speech} & \multicolumn{2}{c}{German} & \multicolumn{2}{c}{French} \\
{} &       EoE &       RMI &       EoE &       RMI &       EoE &       RMI &       EoE &       RMI \\
\midrule
Single   &  80.7±1.5 &         - &  73.7±1.2 &         - &  32.8±2.8 &         - &  27.1±6.3 &         - \\
Ensemble &      83.7 &      97.4 &      67.8 &      73.7 &      39.5 &      82.4 &      25.0 &      73.6 \\
EnD      &  79.5±1.1 &         - &  75.9±0.6 &         - &  35.4±1.6 &         - &  15.6±3.2 &         - \\
\Endd    &  78.3±1.6 &  97.1±0.3 &  77.0±0.3 &  78.5±0.2 &  38.3±1.6 &  70.9±0.7 &  15.9±3.0 &  60.1±3.6 \\
\bottomrule
\end{tabular}
\end{table}

Table~\ref{tab:wmt_pred} further confirms the findings made in the previous section: \Endd via Dirichlet-Proxy outperforms regular ensemble distillation in terms of translation quality and sequence-level error detection. Furthermore, in Table~\ref{tab:wmt_ood} we see that, compared to image classification, the OOD performance gap between total uncertainty and knowledge uncertainty is significantly larger. This might be explained by a significantly larger output space (40,000 classes instead of 1000) or the sequential nature of NMT predictions: because the model generates candidates in a large output space of all possible sequences, its prediction entropy might be high regardless of presence of a domain shift.



\section{Conclusion}\label{sec:conclusion}

This work examined poor convergence of Ensemble Distribution Distillation when applied to large-scale tasks where the number of classes is very high. We investigated the Dirichlet log-likelihood loss and showed that classes with low probability induce larger gradients than high-probability classes, forcing the model to focus on the distribution of the ensemble tail-class probabilities. We proposed a new training objective which minimizes the reverse KL-divergence to a \emph{Proxy-Dirichlet} target derived from the ensemble. This loss resolves the gradient issues of Ensemble Distribution Distillation, as we demonstrate both theoretically and empirically on the ImageNet and WMT17 En-De datasets containing 1000 and 40,000 classes, respectively. This, this work allows Ensemble-Distribution Distillation to be applied to tasks with arbitrary numbers of classes and complexity, enabling fast ensemble inference through distillation in compute bound, risk-critical applications.

\bibliographystyle{IEEEbib}
\bibliography{bibliography}

\end{document}